\pdfoutput=1 
\PassOptionsToPackage{table}{xcolor}
\documentclass[sigconf]{acmart}

\usepackage{epsfig,endnotes}
\usepackage{subcaption}
\usepackage{graphicx}
\usepackage{amsmath}
\usepackage{float}
\usepackage[font=small,labelfont=bf]{caption}
\usepackage{wrapfig}
\usepackage{url}
\usepackage{hyperref}

\usepackage{breakurl}
\usepackage{multirow, multicol, booktabs}
\usepackage[labelfont=sc]{caption}
\usepackage{graphbox}
\usepackage{lipsum}
\usepackage[linesnumbered,algoruled,boxed,lined]{algorithm2e}
\usepackage{algpseudocode}
\acmYear{2024}\copyrightyear{2024}
\setcopyright{rightsretained}
\acmConference[MOBISYS '24]{The 22nd Annual International Conference on Mobile Systems, Applications and Services}{June 3--7, 2024}{Minato-ku, Tokyo, Japan}
\acmBooktitle{The 22nd Annual International Conference on Mobile Systems, Applications and Services (MOBISYS '24), June 3--7, 2024, Minato-ku, Tokyo, Japan}
\acmDOI{10.1145/3643832.3661871}
\acmISBN{979-8-4007-0581-6/24/06}

\begin{document}

\newcommand{\name}{Radarize}
\newcommand{\para}[1]{\vspace{4pt}\noindent\textbf{#1}}
\newcommand{\squishlist}
{
    \begin{list}{$\bullet$}
    {
        \setlength{\itemsep}{0pt}      \setlength{\parsep}{3pt}
        \setlength{\topsep}{3pt}       \setlength{\partopsep}{0pt}
        \setlength{\leftmargin}{1.5em} \setlength{\labelwidth}{1em}
        \setlength{\labelsep}{0.5em}
    }
}
\newcommand{\squishend}
{
    \end{list}
}

\newcommand{\dv}[1]{\textcolor{red}{DV: #1}}
\setlength{\textfloatsep}{10pt plus 1.0pt minus 2.0pt}

\date{}

\title[\name: Enhancing Radar SLAM with Generalizable Doppler-Based Odometry]{\name: Enhancing Radar SLAM with Generalizable \\Doppler-Based Odometry}

\begin{teaserfigure}
    \vspace{-0.15in}
    \begin{center}
    \href{https://radarize.github.io}{\huge\color{blue}\texttt{https://radarize.github.io}}
    \end{center}
\end{teaserfigure}

\author{Emerson Sie}
\affiliation{
    \institution{University of Illinois Urbana-Champaign}
    \country{}
    \city{}
}
\author{Xinyu Wu}
\affiliation{
    \institution{University of Illinois Urbana-Champaign}
    \country{}
    \city{}
}
\author{Heyu Guo}
\affiliation{
    \institution{Peking University}
    \country{}
    \city{}
}
\author{Deepak Vasisht}
\affiliation{
    \institution{University of Illinois Urbana-Champaign}
    \country{}
    \city{}
}

\begin{abstract}

Millimeter-wave (mmWave) radar is increasingly being considered as an alternative to optical sensors for robotic primitives like simultaneous localization and mapping (SLAM). While mmWave radar overcomes some limitations of optical sensors, such as occlusions, poor lighting conditions, and privacy concerns, it also faces unique challenges, such as missed obstacles due to specular reflections or fake objects due to multipath. To address these challenges, we propose \name, a self-contained SLAM pipeline that uses only a commodity single-chip mmWave radar. Our radar-native approach uses techniques such as Doppler shift-based odometry and multipath artifact suppression to improve performance. We evaluate our method on a large dataset of 146 trajectories spanning 4 buildings and mounted on 3 different platforms, totaling approximately 4.7 Km of travel distance. Our results show that our method outperforms state-of-the-art radar and radar-inertial approaches by approximately 5x in terms of odometry and 8x in terms of end-to-end SLAM, as measured by absolute trajectory error (ATE), without the need for additional sensors such as IMUs or wheel encoders.
\end{abstract}



\maketitle

\thispagestyle{empty}

\section{Introduction}
\label{sec:intro}

Simultaneous localization and mapping (SLAM) is a core requisite for many robotics applications. Using SLAM, a robot can simultaneously construct a map of its surrounding environment and localize itself within this map. Traditional SLAM relies on optical sensors (i.e. cameras and LIDAR). However vision-based SLAM suffers from failures in low lighting conditions and raises privacy concerns during indoor use, while LIDAR-based SLAM is thwarted by common obscurants like fog and smoke.

mmWave radars avoid many of these challenges. They can operate in low lighting and occluded settings, e.g., in agricultural robots or search and rescue robots, mmWave radars can continue to operate despite being occluded by dirt and smoke. For indoor robots such as vacuum cleaning robots, radars can be easily hidden behind a facade to not appear prying to users. Moreover, radar has lower resolution than optical sensors, increasing privacy-friendliness. As a result, there is an increasing interest in developing SLAM techniques using frequency-modulated carrier wave (FMCW) mmWave radar \cite{radarhd, millimap}.

\subsection{Challenges for Radar-Based SLAM}\label{sec:challenges}

\begin{figure}[t]
    \centering
    \includegraphics[width=\linewidth]{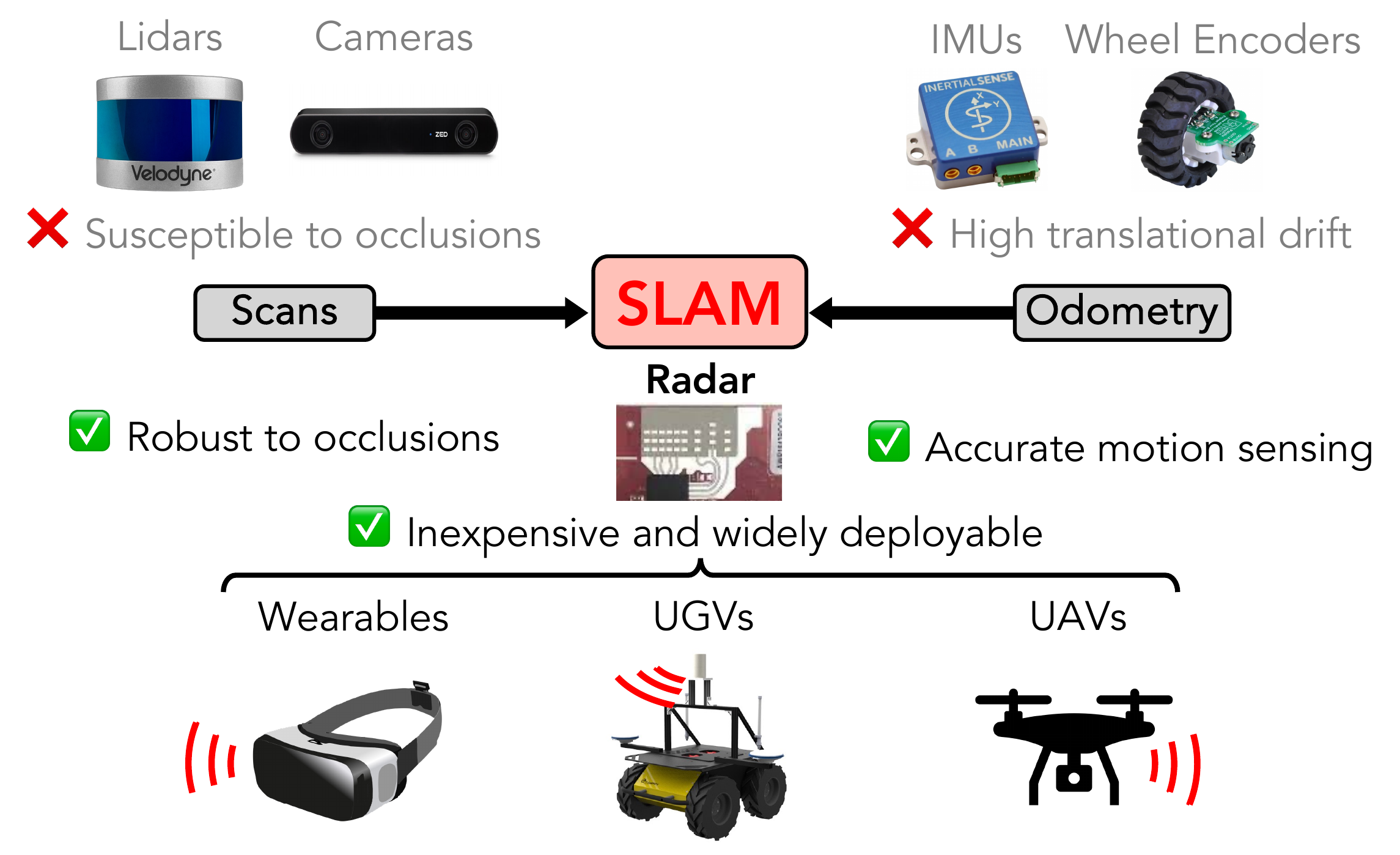}\vspace{-0.2in}
    \caption{Key advantages of radar over conventional SLAM sensors. \name\ leverages both environmental sensing and motion sensing capabilites of radar to perform SLAM.}
    \label{fig:intro}
    \vspace{-0.1in}
\end{figure}

\begin{figure*}[ht!]
    \centering
    \includegraphics[width=\linewidth]{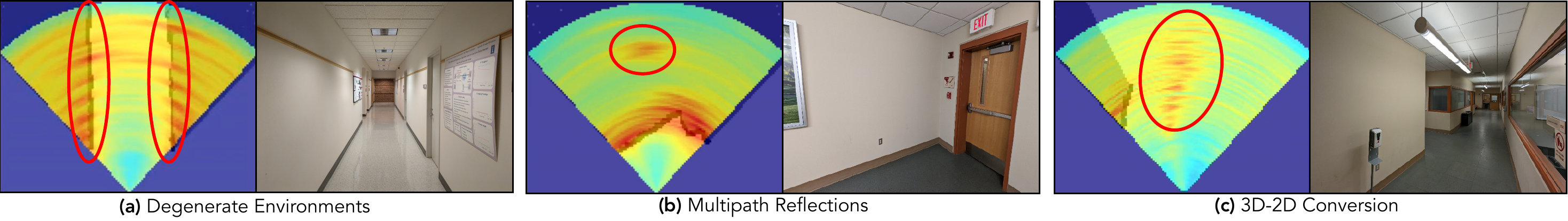}\vspace{-0.15in}
    \caption{Common sources of artifacts in indoor environments. Each image of a scene is paired with a top-down radar heatmap superimposed on a depth camera point cloud. (a) When moving down hallways, featureless flat walls present ambiguities during scan matching. (b) Multipath reflections can cause objects to appear behind surfaces. (c) Limited elevation resolution on most sensors induces artifacts from floors/ceilings.}
    \label{fig:challenges}\vspace{-0.1in}
\end{figure*}

Current techniques for mmWave radar-based SLAM are inadequate due to three key challenges:

    \para{(i) Degenerate Scan Matches: }Past work like RadarHD~\cite{radarhd} uses scan matching techniques to infer a radar's motion. This technique is inherited from optical SLAM pipelines that compare two frames from a camera or LIDAR to estimate the robot's self-motion. We observe that relying on scan matching can induce failure modes for radar-based SLAM similar to those of LIDAR-based SLAM \cite{lim_adalio_2023, filip_lidar_2023}. Specifically, in corridor and hallway-like environments, the lack of distinct geometric features (e.g., due to flat walls) leads to degenerate cases where one frame can be matched with several other frames along the corridor, leading to large errors. To illustrate this, consider Fig. \ref{fig:challenges}(a), which depicts a common hallway environment. We can see that the radar heatmap consists of reflectors along two parallel flat walls (circled in red). If the observer moves down the hallway (i.e. forward) the heatmap after motion would be similar to the heatmap captured just before the motion took place. Hence, during scan matching, the result would be as if no motion occurred. %
    
    \para{(ii) Inadequacy of Inertial Sensors: } A common approach to deal with degenerate environments is to use additional sensors such as accelerometers (e.g., milliMap~\cite{millimap}, milliEgo~\cite{lu_milliego_2020}). This approach carries several disadvantages. First, accelerometers suffer from integrative errors because estimating position from acceleration requires double integration. Second, accelerometers fare badly in scenarios when the robot moves at a near constant velocity which is common in the case of ground vehicles \cite{lee_visual-inertial-wheel_2020}. In such scenarios, the acceleration is nearly unobservable due to being dominated by noise \cite{lv_observability-aware_2022, lee_visual-inertial-wheel_2020, yang_online_2020, yang_d3vo_2020}. Finally it is well known that, in practice, effectively using multiple sensors in conjunction requires laborious precise calibration \cite{yang_online_2020}. Such factory calibration procedures often necessitate expensive bespoke infrastructure (i.e. synchronized clocks, precision turntables, robotic manipulators, motion tracking systems, fiducial targets \cite{xiao_online_2019, li_spatiotemporal_2022, lv_observability-aware_2022}), rendering such procedures out of reach for low-cost and self-assembled sensor suites in the field \cite{yang_online_2020, huang_online_2020}. Furthermore, this calibration may need to be re-performed periodically as the operating conditions of the robot may change over time due to factors such as wear and tear \cite{feng_online_2019, yang_online_2020}.
    
    \para{(iii) Artifacts due to Multipath:}  Radar-based SLAM is prone to two types of artifacts in the generated maps. (a) Multipath effect: radio signal reflections from every day objects cause false images to appear in maps produced by radar-based SLAM approaches. For example in Fig. \ref{fig:challenges}(b), a phantom caused by reflections is seen behind a cul-de-sac. Recent work in radar-based SLAM has not addressed multipath effects~\cite{radarhd, millimap}. (b) 3D-to-2D conversion artifacts: since SLAM focuses on creating a 2D map of the 3D world, past work largely utilizes azimuthal beamforming techniques on single chip mmWave radars. This approach leads to floor and ceiling reflections being mapped to artificial obstacles in the generated map, as shown in Fig.~\ref{fig:challenges}(c). A long ceiling lamp appears in the heatmap even as the radar is pointed horizontally.

\subsection{\name -- Radar-Native SLAM}

We present \name, a novel FMCW radar-based SLAM system that counters these challenges. In designing \name, we take a radar-native approach, i.e., we use properties unique to radio signals that have been underexplored in past work.  Notably, \name\ is self-contained --- it performs SLAM using only a simple off-the-shelf single-chip radar \cite{iwr1843} and does not rely on additional sensors or custom hardware. \name\ is built on the following technical contributions.

\para{Doppler-based Translation Estimation:} Scan-matching algorithms and inertial sensors ignore a key inherent advantage of radio signals -- doppler shift. As a robot explores the environment, the radar signal undergoes doppler shift directly proportional to its velocity. Since doppler shift directly measures velocity, it can accurately estimate translation even in repetitive environments. Furthermore, velocity-to-distance estimates experience lower integration errors and drift compared to accelerometers. Therefore, \name\ utilizes doppler shift as its primary translation estimate. 

Using doppler shift to estimate translation is challenging in indoor settings. In outdoor settings, doppler shift is commonly used to estimate velocities, e.g., to detect speeding cars. Unlike outdoor scenarios that have few big reflectors, indoor settings suffer from a clutter of small reflectors. When the radar moves indoors, it observes reflections from multiple reflectors, each with a different relative velocity. The angle between the reflector and the radar determines the doppler shift experienced by the radar. %
As such, a radar must infer its translation by comparing reflectors from different directions.  \name\ creates doppler-azimuth heatmaps from raw radar signals for this objective, i.e., it represents radar reflections as a function of doppler shifts along different angles. In this doppler-azimuth representation, doppler shift across different angles exhibits a unique signature that captures a radar's self-velocity and heading direction.

\para{Correlation-based Rotation Estimation:} We observe that while doppler shift is accurate for translational motion, it does not capture rotational motion accurately. Therefore, we need to find an alternative mechanism to estimate rotation of the radar. To estimate rotational motion, we follow a two-step approach. First, \name\ performs a coarse grained rotation estimation using two time-shifted radar frames. Specifically, it compares the rotational shift between two range-azimuth heatmaps that demonstrate the reflected signal intensity at different locations in space. We design a neural network with a unique data augmentation procedure and a cyclic consistency loss to accurately estimate the rotation. While this process is similar to scan matching algorithms discussed above, rotational motion is far less likely to suffer from the degenerate scan matching cases (unlike translational motions). In the second stage, \name\ relies on the SLAM pipeline discussed below to fine tune these estimates in conjunction with the map of the environment and the translational motion estimated using doppler shift. 

\vspace{3pt}\noindent\textbf{Artifact Rejection:} To address multipath, \name\ includes a smart multipath rejection scheme. It analyzes the range-azimuth output of the radar, and rejects all but the first reflection, i.e., corresponding to the shortest distance, along each direction. Since multipath reflections travel longer distances, this approach suppresses multipath effect. %
In addition, to counter reflections from floors and ceilings, we introduce an antenna array pre-processing step to reduce reflections from high or low elevations, and a post-processing step to reject such false objects from 2D maps. 

We implemented \name\ using PyTorch \cite{paszke_pytorch_2019} and evaluated it on a large dataset of 146 trajectories collected across 4 different buildings and mounted on 3 different platforms totalling approximately 4.7 Km of travel distance. In terms of odometry, we find that it outperforms competing radar-inertial and neural-inertial approaches by $5 \times$ in terms of absolute trajectory error (ATE). In terms of end-to-end SLAM, we find that this value increases up to $8 \times$. Additionally, our qualitative results (Fig. \ref{fig:showcase_slam}) show that \name\ can enable very long loop closures up to 80m of total travel distance.

\subsection{Summary of Contributions}
Our key contributions are:
\squishlist
    \item We present \name -- a novel SLAM pipeline designed for use with small commodity radar sensors. \name\ exploits the properties of FMCW radar signals to perform radar-based SLAM without need for auxiliary sensors such as IMUs. Yet, \name\ exceeds the performance of radar-based methods leveraging IMUs. %
    \item We identify effective representations of FMCW radar signals for odometry and mapping tasks. We identify sources of artifacts in such representations, and describe techniques for ameliorating them. We also show how data augmentation techniques can enable learning-based techniques to efficiently utilize such representations.
    \item We show via a large scale and comprehensive evaluation that \name\ generalizes across platforms and environments. It works as well when deployed on humans (e.g., as a wearable) as it does on wheeled ground robots, and can map buildings of diverse architectural styles.
    \item We contribute a new dataset for radar-driven SLAM spanning three platforms, four buildings, and 4.7 Km of moving distance\footnote{{Demos, code, and data are available at \href{https://radarize.github.io}{\color{blue}\texttt{https://radarize.github.io}}}}.
\squishend

\section{Background}\label{sec:background}
\noindent\textbf{Radar-based Sensing: }mmWave-based radars have increasingly become accessible and mainstream over the last decade~\cite{felic2015single, lien2016soli}. These devices transmit low power radio signals and capture the reflections of these signals from the surrounding environment. By analyzing these reflections, the radar estimates the distance, angle, and doppler shift corresponding to each reflector in the surrounding.

\noindent\textit{Distance: }A typical radar sensor relies on Frequency Modulated Carrier Wave (FMCW) to measure distance. FMCW signals comprise of chirps, a specific radio signal where the frequency of the signal varies linearly with time. When a reflection is captured by the radar, it is a delayed copy of the chirp and hence lags in frequency. By analyzing the difference in the frequency between transmitted and received chirps, a radar can compute the time delay between them. The time delay corresponds to distance because radio waves travel at the speed of light. The range resolution increases with the bandwidth of the chirp.

\noindent\textit{Angle: } A radar sensor relies on multiple transmit/receive antennas to measure the angle of each reflection. The phase of the received (or transmitted) signal at different antennas depends on the angle of arrival (or departure). By estimating the phase difference across antennas, a radar can identify the corresponding angles. The antenna separation is determined by the signal wavelength, and the antenna span defines the angular resolution (higer span offers better resolution). 

\noindent\textit{Doppler Shift: }A radar can capture multiple reflections over time to estimate the doppler shift corresponding to each reflector in the surrounding. The doppler shift corresponds to the velocity of the reflector. The resolution increases with the total time of signal capture. For more concrete mathematical discussions on radar processing, we refer the reader to~\cite{ti_tutorial}. 

\begin{figure}
    \centering
    \includegraphics[width=\columnwidth]{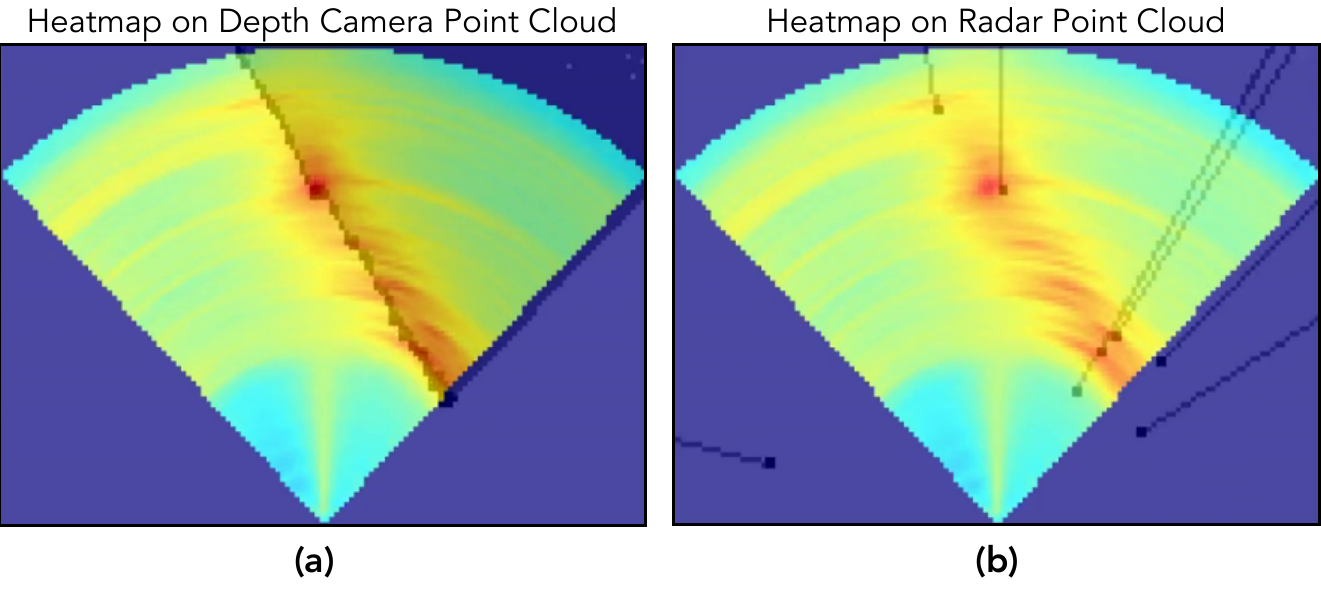}\vspace{-0.1in}
    \caption{%
    The range-azimuth heatmap captures the reflections along different distances and angles. The heatmap is mapped to a cartesian plane for easy visualization. The intensity of reflection varies from blue (low) to red (high). \textit{Left.} Dense heatmap overlaid on depth camera point cloud. \textit{Right.} Same heatmap overlaid on sparse radar point cloud. We trace the shadow behind each point for visibility. 
    }
    \label{fig:weak_reflections}\vspace{-0.1in}
\end{figure}
We will use two different representations of the radar data in this paper. First, we use doppler-azimuth heatmaps to denote a 2-dimensional image where each pixel corresponds to a fixed doppler and angle value. The color of the pixel denotes the intensity of reflection emerging from that doppler and angle value. Similarly, we use range-azimuth heatmaps to demonstrate the reflections emerging at particular range and angle values. For this heatmap, the reflectors at different doppler shifts are squished into a single point.

\para{Simultaneous Localization and Mapping  (SLAM):} In SLAM, a moving agent, such as a robot, creates a map of the environment and estimates its trajectory simultaneously. SLAM relies on two types of inputs: scans (e.g., LIDAR scans) and odometry inputs (e.g., inertial sensors that provide relative motion of the agent between scans). The scans of the environment can come from different sensor types such as LIDARs~\cite{shan_lego-loam_2018,shan_lio-sam_2020}, cameras~\cite{mur-artal_orb-slam_2015,mur-artal_orb-slam2_2017,campos_orb-slam3_2021,qin_vins-mono_2018}, acoustic sensors~\cite{evers_acoustic_2018, dokmanic_acoustic_2016}. These scans are typically represented in the form of probabilistic occupancy grids, where the probability value corresponding to each grid point indicates the probability of presence of an obstacle at that grid point. 

We rely on the Cartographer~\cite{cartographer} framework for 2D-SLAM. Cartographer was originally designed for LIDAR-based SLAM. It uses a small sequence of scans to create local submaps. Within each submap, it matches the current scan to the submap to estimate the pose of the agent. The odometry input serves as a prior for the agent's pose, and isn't necessarily required since scan matching can produce odometry estimates, especially when using dense optical sensors like LIDARs. As the agent walks the environment, the drift in its pose increases over time. To reduce such drift, Cartographer leverages global optimizations using constraints like loop closures, e.g., if a robot revisits the same location, Cartographer can use this scan to correct for accumulated drift. We refer the reader to ~\cite{cartographer} for a detailed discussion.

\section{\name\ Design}
\label{sec:design}

\begin{figure*}[t]
    \centering
    \includegraphics[align=c, width=\linewidth]{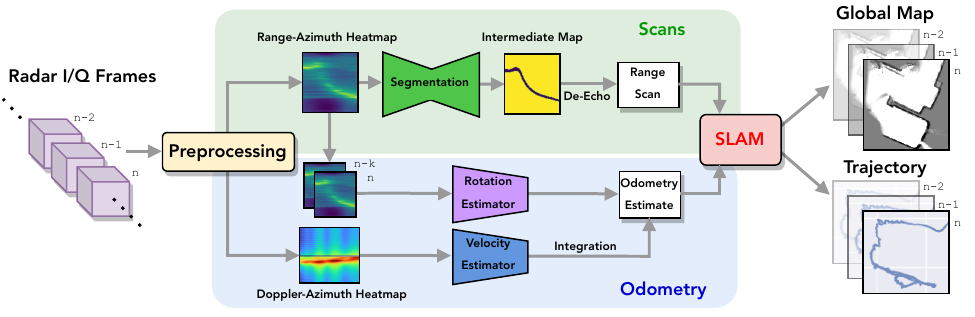}\vspace{-0.1in}
    \caption{\textbf{\name\ Overview.} \textit{Left.} Radar frames are processed into heatmaps. \textit{Top. } The mapping module converts range-azimuth heatmaps into range scans. \textit{Bottom.} The tracking module (a) regresses relative rotation from successive range-azimuth heatmaps and (b) regresses velocities from doppler-azimuth heatmaps and integrates them into odometry estimates. \textit{Right.} An optimization-based SLAM backend outputs real-time map and trajectory estimates. }
    \label{fig:system}\vspace{-0.1in}
\end{figure*}

\begin{figure}[t]
    \centering
    \includegraphics[width=\columnwidth]{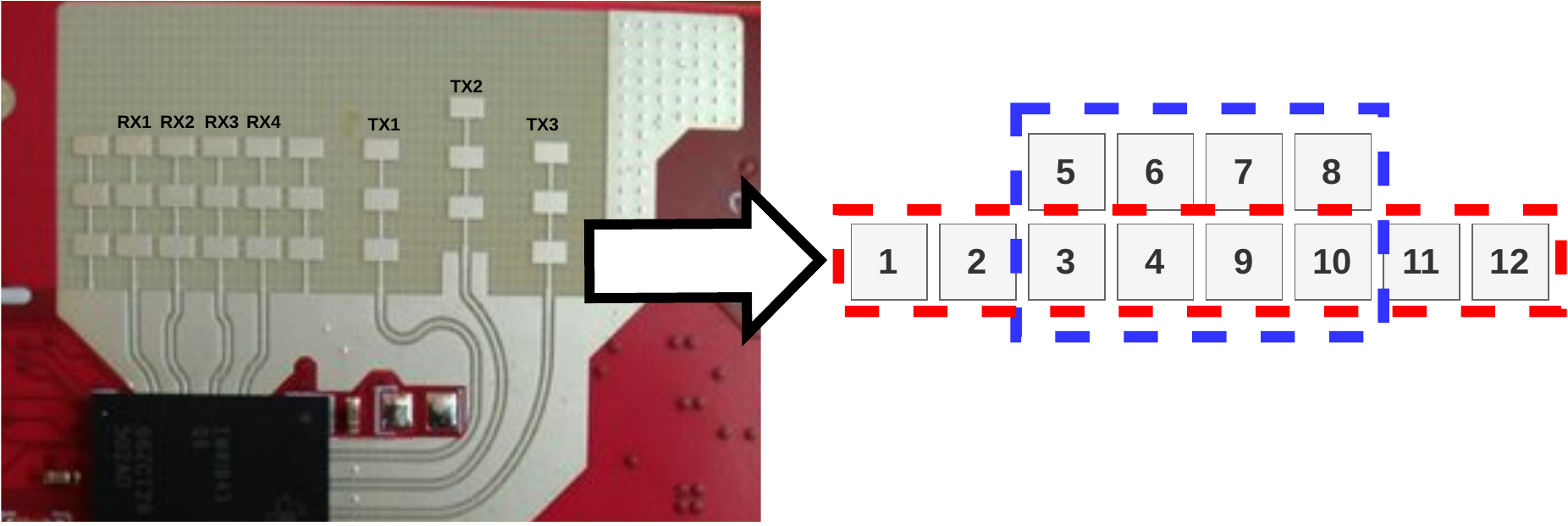}\vspace{-0.1in}
    \caption{\textit{Left.} Physical IWR1843 antenna array. \textit{Right.} Corresponding virtual antenna array. Azimuth-only heatmaps are derived from subarray enclosed in red. Elevation-aware heatmaps are derived from the subarray enclosed in blue.}
    \label{fig:1843_antennas}
    \vspace{-0.2in}
\end{figure}

\begin{figure*}[t]
    \centering
    \includegraphics[width=1.0\linewidth]{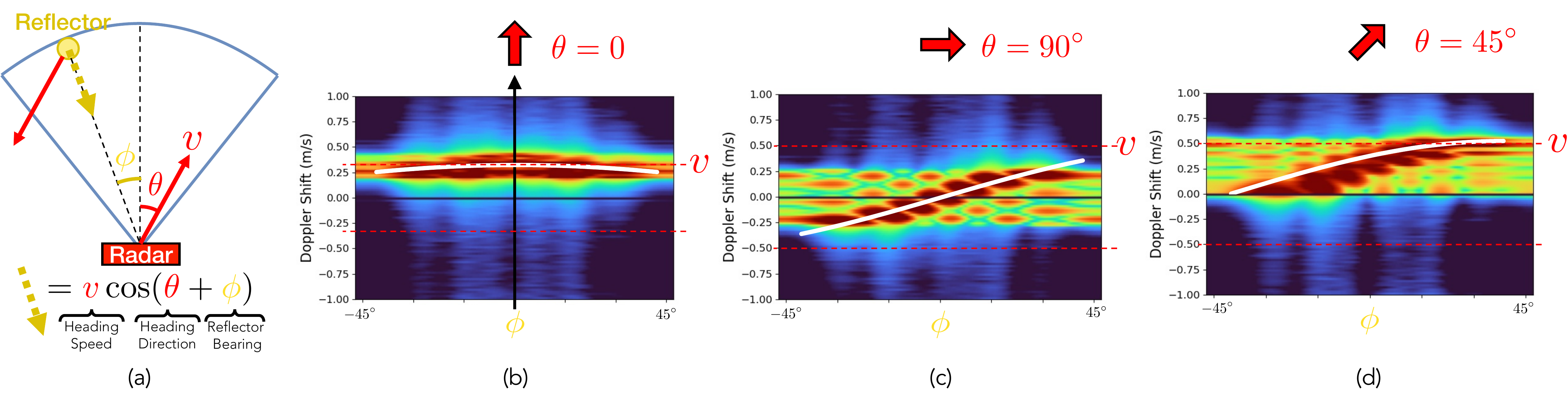}\vspace{-0.2in}
    \caption{Estimating 2D translational motion from doppler-azimuth heatmaps. (a) A forward facing radar "squishes" reflectors in the environment onto the horizontal plane (blue) and measures their velocity and bearing angle $\phi$. (b) Doppler-azimuth heatmap when moving forward at 0.3 m/s. (c) Doppler-azimuth heatmap when moving right at 0.5 m/s. (d) Doppler-azimuth heatmap when moving diagonally (forward + right) at 0.5 m/s. For (b-d), theoretical values are overlaid on the heatmaps as white lines.}
    \label{fig:translation}\vspace{-0.1in}
\end{figure*}

Before we delve deeper into \name, we present a set of goals and choices that motivate our design:

\para{(i) Streamlined Hardware Requirements:} There is a body of past work that relies on large expensive scanning radars~\cite{cen_precise_2018,park_pharao_2020,hong_radarslam_2020, monaco_radarodo_2020} for radar applications. Such radars rely on mechanical motors to spin them and scan their surroundings. We do not consider such radars in this work because they can't support lightweight, compact robots. Instead, we focus on the emerging context of single-chip commodity off-the-shelf (COTS) radars~\cite{iwr1843,awr2944,iwr6843isk}. These radar sensors are inexpensive and small enough to be integrated into mobile devices and small robots. We recognize that more complex radar designs are possible, but we aim for widespread availability, low cost, and ease of use. This choice is consistent with recent work in radar-based SLAM~\cite{sie2023batmobility, radarhd, millimap}.

\para{(ii) Radar-only Design:} Instead of relying on auxiliary sensors like wheel encoders or IMUs, we opt to use radar's intrinsic properties as the source of odometry. This makes sense for two reasons. First, we observe that the intrinsic properties of radar make IMUs redundant for 2D SLAM. Radar measures doppler shift as a first order primitive. Doppler shift corresponds to velocity and as such, is less prone to drift. Hence, the use of doppler shift can effectively supersede accelerometers. Similarly, rotation can be observed from the shifting of reflectors in the environment over time. Second, adding more sensors adds unnecessary complexity and overhead, often necessitating nontrivial calibration procedures and/or synchronization procedures \cite{zhu_robust_2022, yang_online_2022}. By using only a single sensor, we sidestep such concerns.

\para{(iii) Dense over Sparse Methods: } Radar SLAM methods can roughly be categorized into two categories \cite{abu-alrub_radar_2023} (a) dense (direct) methods that rely on dense heatmaps,%
and (b) sparse (indirect) methods that process radar signals into a sparser representation (i.e. keypoints, point clouds) before being used. We show a comparison of a radar heatmap v.s. a radar point cloud in Fig.~\ref{fig:weak_reflections}. As shown, sparse representations lose potentially useful information but offer the advantage of smaller computational complexity and better noise/artifact resilience. In contrast, dense methods preserve all the information at the cost of potential artifacts. We choose the latter approach, opting to capitalize on the benefits of dense methods while introducing new methods to deal with artifacts.

\subsection{Design Outline}

An overview of \name\ is shown in Fig. \ref{fig:system}. \name\ consists of three parts. First, a preprocessing module (shown in beige) converts radar I/Q frames into heatmaps. The tracking module (shown in blue) identifies translation (Sec.~\ref{sec:translation}) and rotation (Sec.~\ref{sec:rotation}) from doppler-azimuth and range-azimuth heatmaps respectively. The mapping module (shown in green) first suppresses vertical reflections (like floors and ceiling), then uses a lightweight segmentation model (i.e. UNet) to segment reflectors on range-azimuth heatmaps (Sec.~\ref{sec:mapping}). The segmentation model is followed by a de-echoing process that performs multipath suppression. Finally, the outputs of the mapping and tracking modules are input to the SLAM backend, which outputs real-time global map and trajectory estimates.

{We tailor our design for the} Texas Instruments IWR 1843 radar~\cite{iwr1843} shown in Fig.~\ref{fig:1843_antennas}(left). This radar has three transmit antennas and four receive antennas. One of the transmit antennas is offset in height. We can transmit using each of the transmit antennas to create a virtual antenna array configuration shown in Fig.~\ref{fig:1843_antennas}(right). We choose this radar because it has good azimuthal angular resolution (8 virtual antennas in the horizontal plane) and is representative of the capabilities of most small radars.

\subsection{Estimating Translation with Doppler-Azimuth Heatmaps}
\label{sec:translation}

Our first goal is to track the radar's egomotion in the horizontal plane. The radar's egomotion consists of 3 degrees of freedom (two translational and one rotational). \name\ factorizes the task of egomotion estimation into translation estimation and rotation estimation. We describe the process of estimating translational motion using a radar's doppler shift below.

Consider the scenario in Fig.~\ref{fig:translation}(a), which depicts a horizontally-pointed forward-facing radar. Suppose the radar undergoes translational motion with speed $v$ along direction $\theta$ (red arrow from the radar). This will induce relative motion from various reflectors in the environment. Suppose there is a reflector along direction $\phi$ (yellow circle). The resultant relative radial velocity (yellow arrow) will be $v\cos(\theta + \phi)$, which in turn induces a doppler shift $\frac{v\cos(\theta+\phi)}{c}f$, where $f$ is the radar signal frequency. 

Our goal is to estimate $v$ and $\theta$ for the robot, given the observed reflections. It is challenging to estimate both $v$ and $\theta$ from a single measurement of doppler shift. However, indoor environments have multiple reflectors. Reflectors are even more abundant at mmWave frequencies, where even smooth surfaces such as walls act as a set of reflectors due to the dispersion effect (see Fig.~\ref{fig:challenges}(a)). While smooth surfaces are often considered a challenge for scan matching (leading to degenerate cases), \name\ can use their abundance of reflections as an advantage. Specifically, given a set of reflectors at different directions, the doppler shift exhibits a unique signature that identifies both $v$ and $\theta$. These signatures present themselves clearly in doppler-azimuth heatmaps, where the doppler shift is plotted as a function of angle $\phi$. %

To illustrate these signatures, we simulate the antenna array depicted in Fig.~\ref{fig:1843_antennas}(red) moving in three velocities (forward at 0.30 m/s, right at 0.50 m/s, forward + right at 0.50 m/s) in a reflector-rich environment, then use beamforming and doppler fast-fourier transform (FFT) to plot the resultant doppler-azimuth heatmaps. The resulting plots are shown in Fig.~\ref{fig:translation}(b--d). For Fig.~\ref{fig:translation}(b), $\theta=0$ and hence, the doppler signature follows the function $v\cos\phi$ and peaks at $\phi=0$ with an amplitude of $v$ (as expected). Similar signatures can be observed for $\theta=\frac{\pi}{2}$ and $\theta=\frac{\pi}{4}$ in Fig.~\ref{fig:translation}(c) and (d) respectively. 
We notice that an intuitive visual structure emerges for different values of $\phi$. The degree of forward (c.f. backward) motion results in increasing upward (c.f. downward) displacement of all the heatmap peaks. The degree of right (c.f. left) motion results in the peaks sloping increasingly positively (c.f. negatively). 

In practice, the heatmaps are sparser and discontinuous, but continue to exhibit similar visual signatures in the doppler-azimuth plots. This interpretable structure motivates us to train a visual model that predicts $v$ and $\theta$ from such doppler-azimuth heatmaps. The details of the model are presented in Sec.~\ref{sec:implementation}. Once we have a model for the velocity, we derive the inter-frame translation by multiplying with the known inter-frame arrival time.

\subsection{Estimating Rotation with Range-Azimuth Heatmaps}\label{sec:rotation}

\begin{figure}[t]
    \centering
    \includegraphics[width=.9\linewidth]{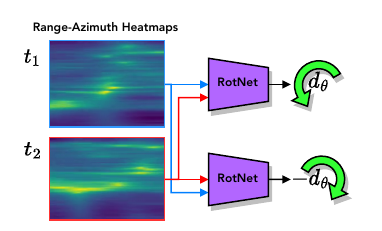}\vspace{-0.1in}
    \caption{Cyclic consistency loss used to train rotation model. The radar undergoes right rotation (and some translation) from $t_1$ to $t_2$, causing the heatmap at $t_2$ to be left-shifted (modulo some warping). }
    \label{fig:rotation}\vspace{-0.1in}
\end{figure}

Using a single-chip radar for rotation estimation is an underexplored and challenging problem. Observe that the doppler-based estimates do not accurately capture rotational motion. This is because during rotational motion, the radial velocity of the reflectors with respect to the radar is zero. Therefore, we cannot rely on our doppler-based translation estimation approach for this task. 

\name's rotation estimation approach relies on range-azimuth heatmaps generated by the radar. Consider the radar in Fig.~\ref{fig:translation}(a) undergoing pure rotational motion e.g. rotating clockwise (c.f counter-clockwise). Then it is expected that the range-azimuth heatmap (see example in Fig.~\ref{fig:rotation}) after rotation will undergo a coherent linear transformation, i.e., all reflectors will shift left (c.f. right) as compared to the version of the range-azimuth heatmap before rotation. This is because under pure rotation, each reflector's range to the radar remains constant, but their bearing angles change by the exact same amount. Moreover, their shift along the azimuth axis is directly proportional to the degree of the radar's rotation. We plan to use this intuition to estimate the rotational motion.

However, precisely matching reflectors across different frames is a challenging task due to three reasons. First, the low resolution of heatmaps generated by low-cost radars makes reflector identification challenging. Second, when translation motion is mixed with rotation, reflectors in range-azimuth heatmaps undergo various degrees of warping. Finally, some reflectors are only visible at certain angles and do not appear in some heatmaps. 

We notice that this task is similar to feature matching in the vision domain. Given the success of machine learning in visual feature matching, \name\ trains a visual model (details in Sec.~\ref{sec:implementation}). Given two successive range-azimuth heatmaps, \name's model estimates the relative rotation between them by measuring the extent of coherent left or right-translation (ignoring some potential incoherent warping due to translational motion) by making appropriate visual correspondences between the heatmaps.

We propose two data augmentation strategies to make the model generalize to various rotation and translation speeds. First, we vary the time spacing between two successive range-azimuth frames, i.e., we vary $t_2-t_1$ in Fig.~\ref{fig:rotation} to expand the range of shifting and warping within the dataset. This helps us augment our training dataset with different velocities. Second, we adopt a cyclic-consistency loss due to the fact that any rotation from $t_1$ to $t_2$ implies the same rotation in reverse from $t_2$ to $t_1$, as depicted in Fig.~\ref{fig:rotation}. We adopt similar strategy at inference time -- we obtain two estimates of rotation by feeding the heatmaps in forward and reverse orders. We average these estimates for the final output.

\subsection{Generating Local Maps}\label{sec:mapping}

\begin{figure}
    \centering
    \includegraphics[width=\linewidth]{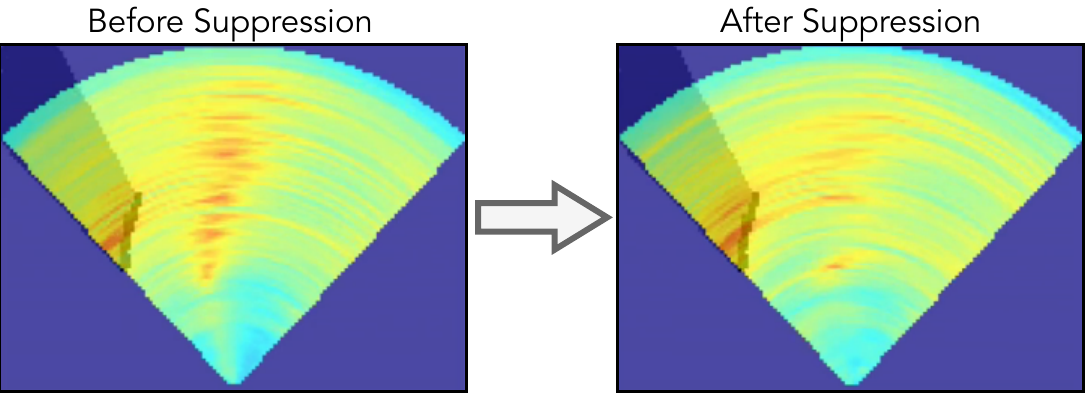}
    \caption{Effect of elevation beamforming on range-azimuth heatmap in Fig. \ref{fig:challenges}(c). \textit{Left.} Original heatmap. \textit{Right.} With elevation beamforming. The artifact is significantly diminished.}
    \label{fig:elevation_awareness}\vspace{-0.1in}
\end{figure}
Our next goal is to generate local maps of the environment surrounding the robot. Multiple local maps are combined together to form the global map using our SLAM backend. As discussed in Sec.~\ref{sec:challenges}, the key challenge for accurate mapping is the artifacts due to  3D-2D conversion artifacts (Fig.~\ref{fig:challenges}(c)) and multipath effects (Fig.~\ref{fig:challenges}(b)).

\para{Artifact Rejection:} Consider the antenna array in Fig.~\ref{fig:1843_antennas} which consists of two virtual arrays -- the 1D azimuth-only antenna array (red) and the 2D elevation-aware antenna array (blue). Deriving a range-azimuth heatmap from the former has a high azimuthal angular resolution. However, it is not elevation-aware, and thus is prone to artifacts from floors and ceilings (Fig.~\ref{fig:elevation_awareness} left). We use the 2D virtual array to suppress such artifacts. Specifically, we create an elevation-aware range-azimuth heatmap by setting the beamforming weights such that the radar points straight ahead in the elevation dimension. This elevation-aware heatmap encapsulates elevation information at the cost of azimuthal angular resolution. We can see an example of such artifact suppression in the heatmap in Fig.~\ref{fig:elevation_awareness} (right). Our observation is that legitimate in-plane reflectors are consistent across both heatmaps, while ceiling and floor reflectors are suppressed in the elevation-aware heatmaps. 

We aim to have \name's mapping module benefit from this observation. Therefore, for each incoming radar frame, we compute both the pure-azimuthal heatmap as well as the elevation-aware heatmap. We then feed both heatmaps to a segmentation model. The model can leverage the former heatmap to output a high resolution map, and leveraging comparisons between the two to weed out 3D to 2D conversion artifacts.

\para{Echo Suppression:}  Another source of errors is the presence of phantoms due to multipath propagation (Fig. \ref{fig:challenges}(b)). We propose a simple heuristic to diminish the effect of such artifacts -- we remove all but the first reflection along each direction from the output of the segmentation module. Intuitively, the output of the segmentation module is a pixel-wise binary classification of the reflectors within a scene. This can be interpreted as a 2D occupancy grid of points, some of which are behind others from the radar's point of view. Due to the presence of phantoms in radar heatmaps, we should place most of our confidence only on the first reflector encountered along any particular bearing angle. %

\subsection{Integrating Odometry and Mapping}
We feed the odometry outputs from our tracking module and the sequence of scans from our mapping module to the state-of-the-art optimization-based SLAM backend (Cartographer~\cite{cartographer}). As a production-grade SLAM system, Cartographer can continuously merge these inputs to generate a map of the environment and the trajectory of the device in real-time, even on computationally lightweight devices. 
\section{Implementation}
\label{sec:implementation}

\subsection{Hardware Design}

\begin{figure}[t]
    \centering
    \includegraphics[width=.7\columnwidth]{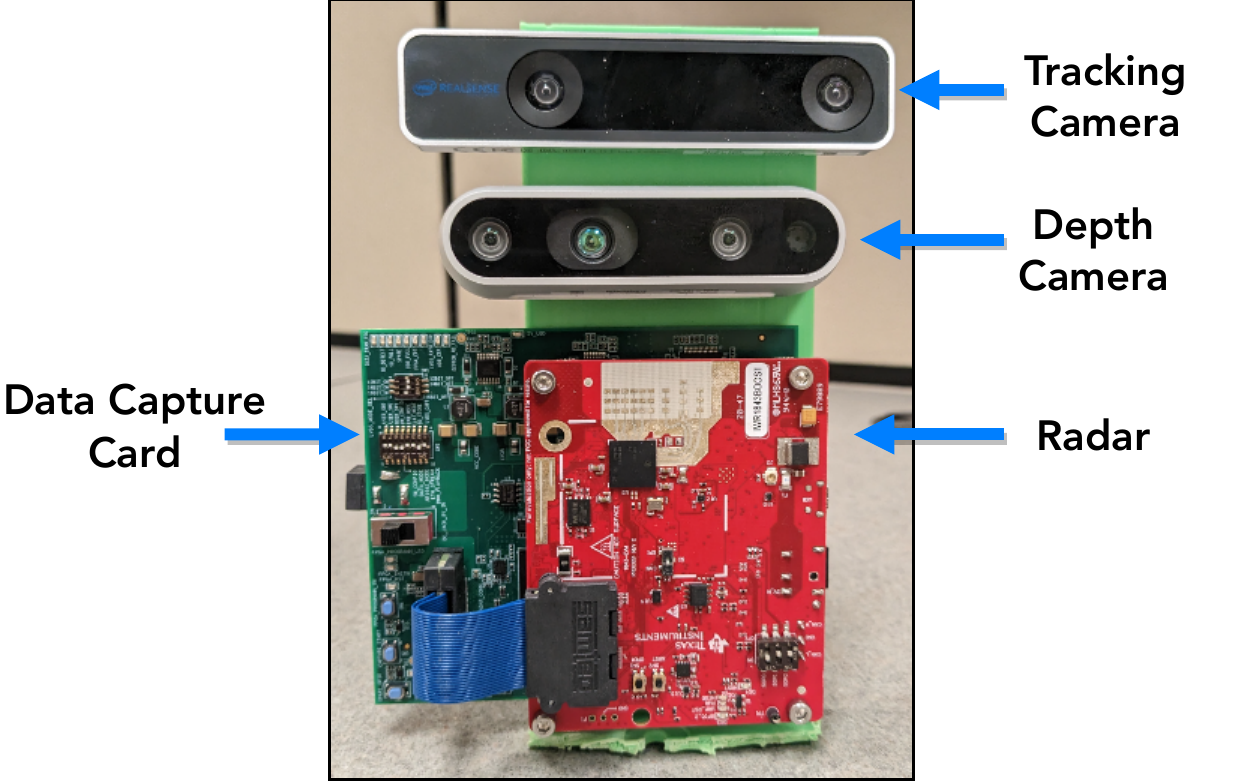}\vspace{-0.1in}
    \caption{Sensor suite used for data collection.}
    \label{fig:hardware}\vspace{-0.1in}
\end{figure}

\noindent\textbf{Sensors:} We use the hardware setup shown in Fig. \ref{fig:hardware}. It consists of an Intel T265 tracking camera \cite{t265}, Intel D435i stereo depth camera \cite{d435}, an Texas Instruments IWR1843 radar \cite{iwr1843}, and a DCA1000 data capture card \cite{dca1000} attached to a 3D printed fixed frame. We use the tracking and depth cameras solely to obtain the pseudo-ground truth trajectories and depth maps respectively. The data capture card is used to capture unprocessed I/Q samples from the radar rather than preprocessed sparse point clouds. %

In total, this setup allows us to capture a sequence of colocated depth images from a stereo depth camera, accelerometer and gyroscope readings from its built-in IMU, sparse point clouds from the radar, frames of radar I/Q samples from the data capture board, as well as pseudo-ground truth poses as obtained from a state-of-the-art visual odometry sensor. \textit{Note that during deployment, only the data capture card and radar components are necessary.}

\para{Parameters:} We tune the parameters of the sensors to be appropriate for indoor settings. We set the chirp parameters of the radar such that the maximum unambiguous range is 4.284m and the range resolution is 4cm. We adjust the maximum depth of the depth camera to match that of the radar. Since the depth camera's horizontal field of view is limited to 88 degrees ($[-44^\circ,44^\circ]$ deviation from boresight), we compute range-azimuth heatmaps within the same range (with an angular bin spacing of $1^\circ$). As a result, the size of our range-azimuth heatmaps is $N_r = 96$ by $N_\theta^r = 88$. 

Additionally, we adjust the inter-chirp interval to be $516\mu s$. This yields $v_{max} = \frac{\lambda}{4*T_c} = 1.89$ m/s, which is a reasonable upper bound on human walking speed. For our doppler-azimuth heatmaps, we find that $N_c = 96$ and $N_{\theta}^c = 180$ (with $1^\circ$ angular bin spacing) to be a good value. This gives us a velocity resolution of $v_{res} = \frac{\lambda}{2*N_c*T_c} = 0.04$ m/s. Finally, the sensors operate at different frequencies (i.e. 200 Hz for the IMU, 100 Hz for the camera). We set the radar frame rate to be 30 Hz and synchronize all sensor readings to the radar.

\subsection{Dataset Collection}

We collect a large and diverse dataset of trajectories spanning several floors across four different campus buildings A, B, C, and D using the set of sensors depicted in Fig. \ref{fig:hardware}.  During data collection, we ensure that there are no other moving objects in the environment. This is consistent with past work in SLAM, which classically assumes a static environment. 

\para{Buildings:} The buildings were chosen to maximize the diversity of architectural styles and construction materials. Building A is a modern construction featuring many glass and metallic surfaces. Building B is the most dated construction and primarily consists of wood and drywall. Building C comprises mainly of brick and metallic surfaces. Building D is an intermediary between building B and C. We choose random floors in each building to randomize the layout. 

\begin{figure}[t]
    \centering
    \includegraphics[width=\columnwidth]{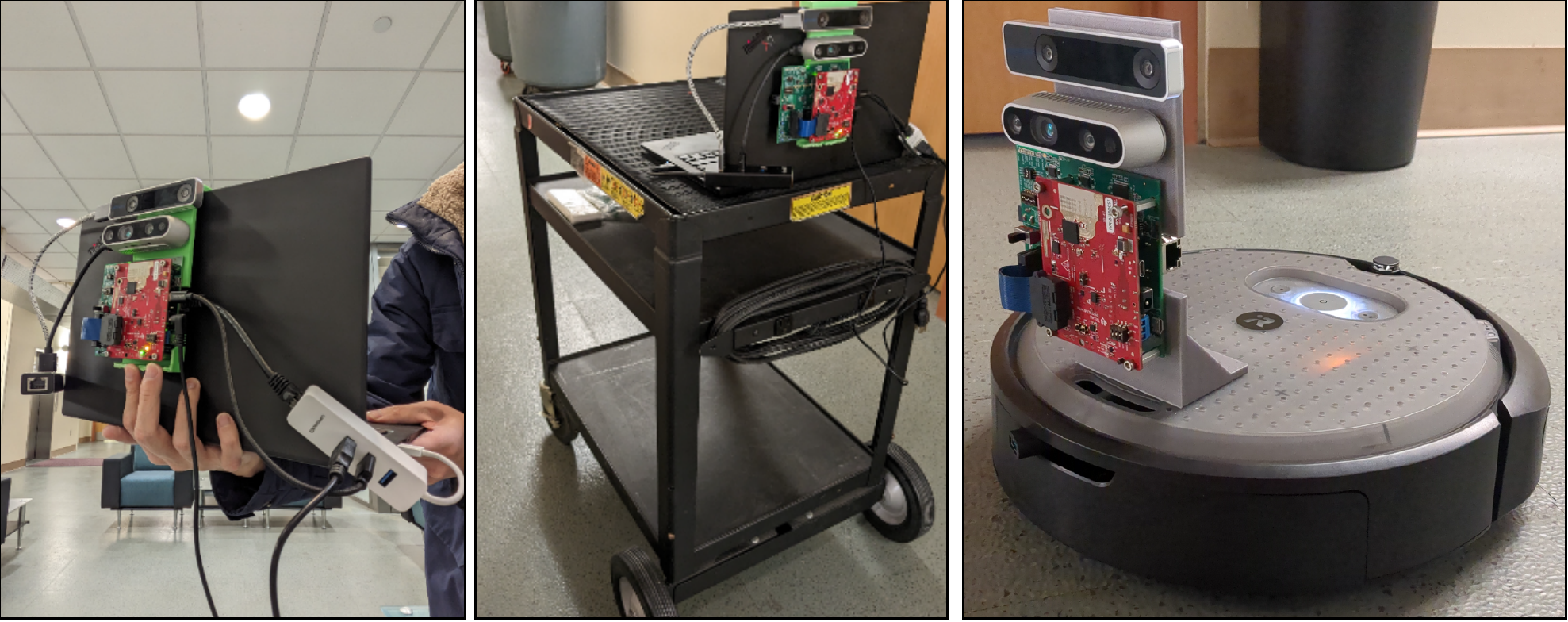}\vspace{-0.1in}
    \caption{\textit{Left.} Moving sensors by hand. \textit{Center.} Sensors mounted on a cart. \textit{Right.} Sensors mounted on robot.}
    \label{fig:data_collection}\vspace{-0.15in}
\end{figure}

\begin{table}[t]
\centering
\rowcolors{2}{}{gray!10}
\begin{tabular}{>{\columncolor{white}}lccccc}
\toprule
\multirow{2}{*}{\textbf{Platform}}                    & \multicolumn{4}{c}{\textbf{Building}} &      \multirow{2}{*}{\textbf{Total}} \\ 
\cmidrule{2-5}
\multicolumn{1}{c}{}                                  & A & B & C & D &  \\ \midrule
Handheld                                              & 15                     & 9                      & 12                     & 11                     & 47\\
Cart                                                  & 10                     & 11                     & 15                     & 9                      & 45\\
Roomba                                                & 9                      & 15                     & 15                     & 15                     & 54\\ 
\midrule
\textbf{Total}                                        & 34            & 35            & 42            & 35            & 146
\end{tabular}
\caption{Breakdown of the trajectories in our dataset.}
\label{tab:dataset}\vspace{-0.3in}
\end{table}

\para{Platforms:} In addition to diversity of environments, we also aim for a diversity of platforms. We move the sensors in one of three ways as shown in Fig. \ref{fig:data_collection}: \textit{(a) Hand-held:} We instruct a set of volunteers to carry the sensors by hand while walking around the building. The volunteers are encouraged to encompass the full range of 2D top-down motions -- for example, they can choose to move backwards or sideways through a narrow hallway. This leads to a challenging set of trajectories featuring sudden accelerations and fast rotations. \textit{(b) Cart-driven:} We mount the sensors onto a cart and push it around the building. The trajectories are more constrained and consist mostly of constant linear velocity motion and slow rotations. Across both aforementioned cases, the set of linear speeds range up to a brisk walking pace of 1.6 m/s. \textit{(c) Robot-driven: }We mount the sensors onto a small ground robot (the iRobot Create3 \cite{create3}) and use teleoperation to drive the robot around the building. The robot has a maximum linear velocity of 0.5 m/s and holds the sensors significantly closer to the ground than the previous cases.

\para{Dataset Summary:} We collect 146 trajectories, with roughly even division across movement types and buildings. Each trajectory consists of around a minute of capture time. The mean travel distance across all handheld and cart-driven trajectories is over 50m. Owing to the relatively slow speed of the ground robot, the mean travel distance for the robot trajectories is around 20m. In total, our dataset consists of roughly 226,800 data samples and 4680m of travel distance. Table \ref{tab:dataset} contains a summary of the dataset.

\para{Caveat on Adverse Conditions:} We do not collect data in adverse conditions (e.g. in occluded environments or low-light environments) for two reasons. First, such conditions would make the optical sensors (i.e. depth and tracking camera) fail and produce junk values, depriving us of the pseudo ground-truth for evaluation. Second, numerous prior works have already shown the robustness of radar to both occluded environments and low-light environments. Specifically, \cite{radarhd} immersed a radar within a smoke chamber and found no difference in the output, and \cite{sie2023batmobility} reports identical performance of radar in both light and dark conditions.

\subsection{Model Architecture \& Training}

\noindent\textbf{Translation Model: } We use a CNN based on ResNet18 \cite{he2016deep}. We modify the input to accept a single-channel doppler-azimuth heatmap of size $N_c \times N_{\theta}^c$ and modify the output layer to output a two-dimensional velocity estimate. To train the model, we use ground truth velocity estimates as obtained from our tracking camera as supervision. We use the standard L2 (mean-squared error) loss for training. Our optimizer is \verb|Adam| \cite{kingma_adam_2014} with parameters $\beta_1 = 0.9$, $\beta_2 = 0.999$. We use a learning rate of $1\mathrm{e}-3$ and a batch size of $128$. For data augmentation, we apply random vertical flipping on the doppler-azimuth heatmap images. Specifically, for a data sample consisting of a doppler-azimuth heatmap image and a velocity label $v$, we vertically flip the doppler-azimuth heatmap and assign the label $-v$. 

\vspace{3pt}\noindent\textbf{Rotation Model: } We again use a CNN based on ResNet18 \cite{he2016deep}. We modify the input to accept a two-channel image of size $N_r \times N_{\theta}^r$ consisting of stacked range-azimuth heatmaps i.e. a range-azimuth heatmap observed at current time, $H_n$, with a previous observation $H_{n-k}$, where $k > 1$. The output is a single value $\hat{\Delta\theta}_{n-k}^n$ corresponding to the rotation estimate. To train our model, we use the pseudo-ground truth rotation estimates from the tracking camera as supervision. We use a standard L2 (mean-squared error) loss with cyclic consistency constraint given by %
\begin{equation}
L_{rot}(H_{n-k},H_n) = \|\Delta\theta_{n-k}^n - \hat{\Delta\theta}_{n-k}^n\|_2^2 + \|\Delta\theta_{n}^{n-k} - \hat{\Delta\theta}_{n}^{n-k}\|_2^2
\end{equation}
Our optimizer is \verb|Adam| \cite{kingma_adam_2014} with parameters $\beta_1 = 0.9$, $\beta_2 = 0.999$. We use a learning rate of $1\mathrm{e}^{-3}$ and a batch size of $128$. We leverage two kinds of data augmentation. First, we design a horizontal flipping data augmentation. Specifically, for an incoming data sample (which consists of a stacked $H_n$ and $H_{n-k}$ and label $\Delta\theta_{n-k}^n$) we randomly apply horizontal flipping on the stacked heatmap image and set the label of the sample to be $-\Delta\theta_{n-k}^n$ accordingly. Secondly, we vary $k$ randomly for each training sample to lie within a small range -- we find empirically that 1-5 frames is a good value.

\vspace{3pt}\noindent\textbf{Segmentation Model: } We choose UNet \cite{navab_u-net_2015} as a backbone as it is fast and achieves state-of-the-art results on segmentation tasks. The input to the model is a two-channel image of size $N_r \times N_{\theta}^r$. This is obtained by stacking the pure-azimuthal and elevation-aware range-azimuth heatmaps. The output of the model is a 2 channel range-azimuth image of occupancy logits of the same size, which we convert into probabilities using the softmax function. Thresholding the output converts it into a 2D grid of discrete points. For the ground truth, we use the point clouds given by the stereo depth camera. For training, we use the Dice loss with \verb|Adam| \cite{kingma_adam_2014}. Our optimizer is \verb|Adam| \cite{kingma_adam_2014} with parameters $\beta_1 = 0.8$, $\beta_2 = 0.9$. We use a learning rate of $1\mathrm{e}-4$ and a batch size of $48$. For data augmentation, we use random horizontal flipping on the range-azimuth heatmaps and depth ground-truth.
\section{Evaluation}
\label{sec:evaluation}

We evaluate our method by contrasting it with several representative state-of-the-art baselines (or combinations thereof):

\squishlist
\item \textbf{RadarHD} \cite{radarhd} is a radar super-resolution method. It leverages an asymmetric UNet architecture to upsample range-azimuth heatmaps into high angular resolution point clouds. The point clouds are subsequently utilized for downstream localization and mapping tasks.
\item \textbf{milliEgo} \cite{lu_milliego_2020} is an radar-inertial odometry estimator based on sparse point clouds. It leverages a cross-attention mechanism to merge odometry estimates from point cloud observations with auxiliary IMU data.
\item \textbf{RNIN} \cite{chen_rnin-vio_2021} is a neural-inertial navigation model that regresses IMU readings into odometry estimates. These estimates are designed to be integrated into various downstream state estimation pipelines.
\squishend

We use the publicly available implementation and pretrained weights for the baselines. For fairness, we also fine-tune the models on our own dataset.

\para{Metrics:} To compute the relevant evaluation metrics we use the \verb|evo| package \cite{grupp2017evo}, widely used to benchmark the performance of odometry and SLAM algorithms. We consider two main metrics: Absolute Trajectory Error (ATE) and Relative Error (RE) for both position and rotation (heading). While ATE captures errors over the entire trajectory, RE captures relative error over smaller trajectory chunks. We refer the reader to \cite{zhang_tutorial_2018} for an in-depth explanation of these metrics.

\subsection{Experimental Setup}

Our experiments are designed to answer the following:

\squishlist
    \item How well does \name's compare with IMU-aided and/or super-resolution based methods?
    \item How well does \name\ generalize across different environments and platforms?
\squishend

To answer the first question, we split our dataset into 5 random training/validation/testing spits in a ratio of 70/10/20 and aggregate the results across all splits. To answer the second question, we consider train/test splits across different buildings and platforms types. 

\subsection{Odometry Performance}

\begin{table}[ht!]
\centering
\rowcolors{3}{gray!10}{}
\begin{tabular}{lccccc}
\toprule
\multicolumn{1}{c}{\multirow{2}{*}{\textbf{Method}}} & \multicolumn{2}{c}{\textbf{ATE}}                            & \multicolumn{2}{c}{\textbf{RE}}                             &  \\ \cmidrule{2-3} \cmidrule{4-5}
\multicolumn{1}{c}{}                                 & \multicolumn{1}{c}{Pos (m)} & \multicolumn{1}{c}{Rot (rad)} & \multicolumn{1}{c}{Pos (m)} & \multicolumn{1}{c}{Rot (rad)} &  \\ \midrule
milliEgo \cite{lu_milliego_2020}                     & 4.100                       & 1.054                         &  0.059                      & 1.018                         &  \\
RNIN \cite{chen_rnin-vio_2021}                       & 2.993                       & 0.542                         &  0.018                      & 0.524                         &  \\
\rowcolor{green!10}
\textbf{\name\ (Ours)}                                                 & \textbf{0.805}              & \textbf{0.366}                &  \textbf{0.014}             & \textbf{0.355}                &  \\ \bottomrule
\end{tabular}
\caption{Comparing odometry performance. Results are averaged across 5 random dataset splits. Lower is better.}
\label{tab:odometry}

\centering
\rowcolors{3}{gray!10}{}
\begin{tabular}{lccccc}
\toprule
\multicolumn{1}{c}{\multirow{2}{*}{\textbf{Method}}}     & \multicolumn{2}{c}{\textbf{ATE}}                            & \multicolumn{2}{c}{\textbf{RE}}                             &  \\ \cmidrule{2-3} \cmidrule{4-5}
\multicolumn{1}{c}{}                                     & \multicolumn{1}{c}{Pos (m)} & \multicolumn{1}{c}{Rot (rad)} & \multicolumn{1}{c}{Pos (m)} & \multicolumn{1}{c}{Rot (rad)} &  \\ \midrule
milliEgo \cite{lu_milliego_2020}                         & 3.977                       &  0.817                        &  0.050                      & 0.807                         &  \\
RNIN \cite{chen_rnin-vio_2021}                           & 5.122                       &  1.927                        &  0.019                      & 1.904                         &  \\
\rowcolor{green!10}
\textbf{\name\ (Ours)}                                                     & \textbf{1.121}              &  \textbf{0.565}               &  \textbf{0.013}             & \textbf{0.559}                &  \\ \bottomrule
\end{tabular}
\caption{Testing odometry generalization to OOD motion. Each method was trained on the handheld trajectories and tested on the cart and robot trajectories. }
\label{tab:odometry_generalization}
\vspace{-0.2in}
\end{table}

\begin{figure*}[t!]
    \centering
    \includegraphics[width=.8\linewidth]{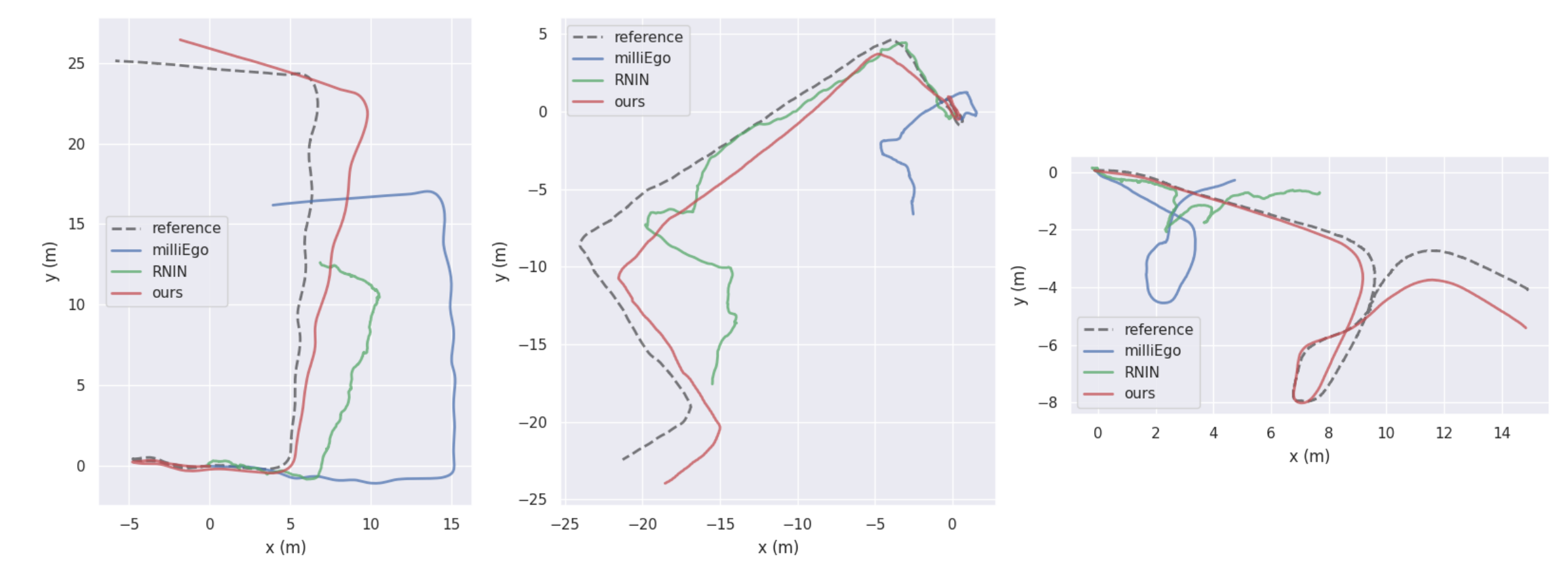}\vspace{-0.2in}
    \caption{\textbf{\name's odometry achieves correct scale.} We observe IMU-based methods are able to reconstruct the shape of long and linear trajectories, but struggle with absolute scale.}
    \label{fig:showcase_odom_a}
\end{figure*}

\begin{figure*}[t!]
    \centering
    \includegraphics[width=.9\linewidth]{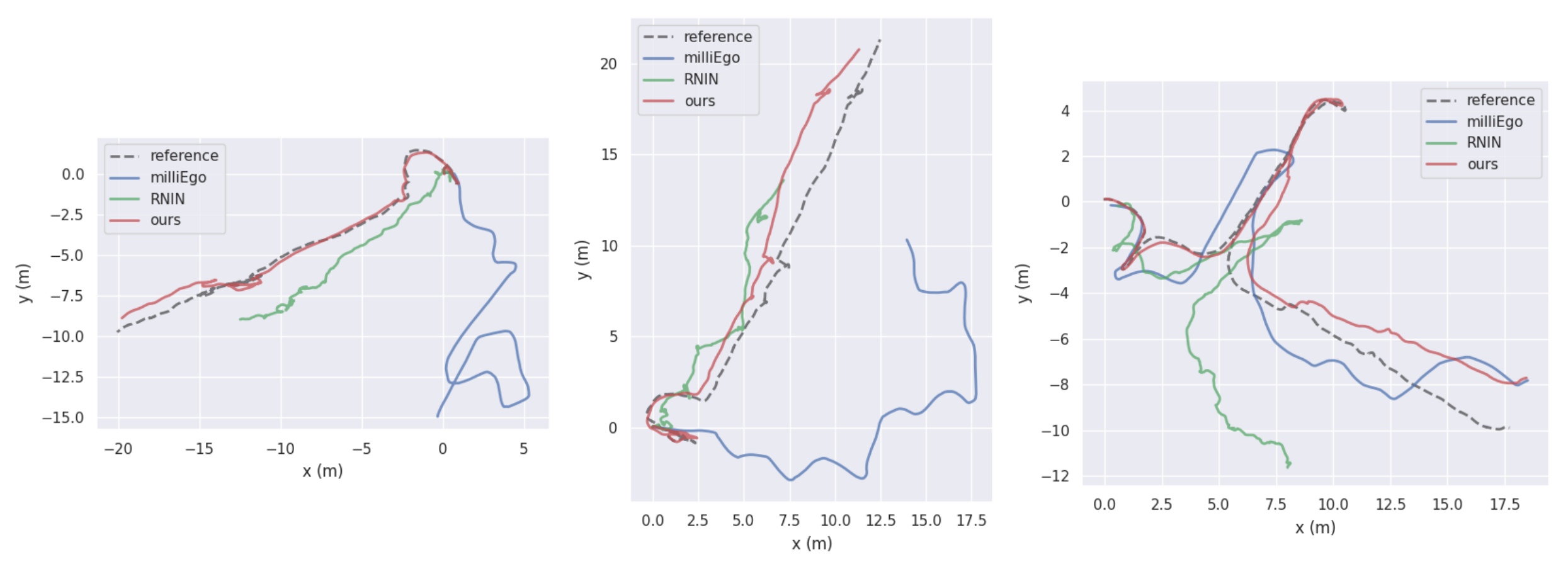}\vspace{-0.2in}
    \caption{\textbf{\name's odometry is flexible.} Our method excels at the handheld trajectories which feature a wide range of motions, whereas the other methods become brittle.}
    \label{fig:showcase_odom_b}
\end{figure*}

\begin{figure*}[t!]
    \centering
    \includegraphics[width=\linewidth]{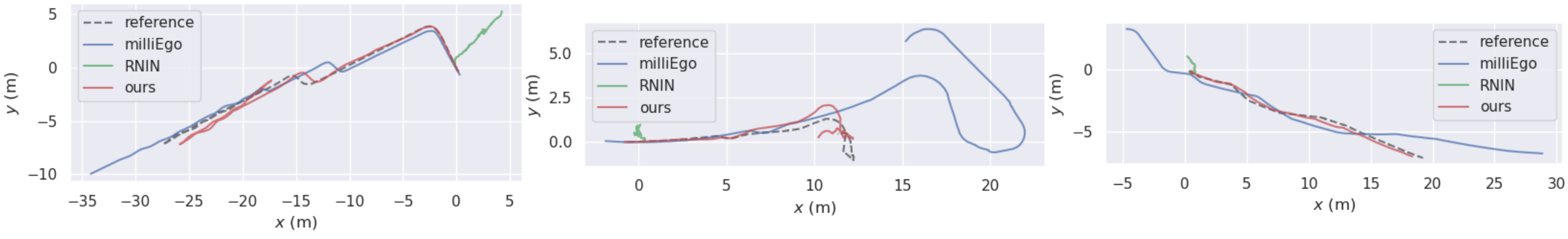}\vspace{-0.2in}
    \caption{\textbf{\name's odometry is robust.} We compare performance on OOD trajectories by training on the handheld trajectories and testing on wheeled ones. Methods leveraging IMUs struggle, whereas \name\ retains similar performance.}
    \label{fig:showcase_odom_c}
\end{figure*}

First, we evaluate our odometry estimation pipeline (Sec. \ref{sec:design}). We summarize our results in Table \ref{tab:odometry}. Our experiments show that our odometry estimation method is superior to the baselines by a wide margin, in spite of not leveraging any IMU information. We present a detailed breakdown below.

\begin{table*}[t!]
\centering
\rowcolors{3}{gray!10}{}
\begin{tabular}{llccccc}
\toprule
\multicolumn{2}{c}{\textbf{Method}} & \multicolumn{2}{c}{\textbf{ATE}} & \multicolumn{2}{c}{\textbf{RE}} &  
\\ \cmidrule{1-2} \cmidrule{3-4} \cmidrule{5-6}
\rowcolor{white} \multicolumn{1}{c}{Odometry} & \multicolumn{1}{c}{Scans} & \multicolumn{1}{c}{Position (m)} & \multicolumn{1}{c}{Rotation (rad)} & \multicolumn{1}{c}{Position (m)} & \multicolumn{1}{c}{Rotation (rad)} &  \\ \midrule
\multicolumn{1}{c}{---} & RadarHD \cite{radarhd}                               & 4.833                       &  0.751                        & 0.045                       & 0.726                         &  \\
milliEgo \cite{lu_milliego_2020} & RadarHD \cite{radarhd}   & 3.900                       &  0.808                        & 0.022                       & 0.780                         &  \\
RNIN \cite{chen_rnin-vio_2021} & RadarHD \cite{radarhd}     & 3.299                       &  0.541                        & 0.020                       & 0.522                         &  \\
\name\ (Ours)   &        RadarHD \cite{radarhd}             & 0.658                       &  0.154                        & 0.018                       & 0.149                         &  \\
\rowcolor{green!10}
\textbf{\name\ (Ours)} & \textbf{\name\ (Ours)}           & \textbf{0.606}              &  \textbf{0.116}               & \textbf{0.017}              & \textbf{0.113}                &  \\ 
\bottomrule
\end{tabular}
\caption{Comparing SLAM performance. Results are averaged across 5 random dataset splits. Lower is better.}
\label{tab:slam}
\vspace{-0.2in}
\end{table*}

\para{Achieving Accurate Scale:} In general, we observe that IMU-based methods can reconstruct the shape of long trajectories, but struggle with its scale. This is substantiated by our cart-driven experiments, which consists primarily of forward motion down long hallways (Fig. \ref{fig:showcase_odom_a}). This is due to the IMU-based methods having to perform double integration for translation but only single integration for heading. As such, IMU-based methods perform better on heading direction estimation than on translation estimation. By contrast, \name\ almost always achieves the correct scale because it directly estimates velocity and needs only single integration for translation.

\para{Accommodating Diverse Motions:} Recall that in our dataset, we collect handheld trajectories with more unrestrained motion than when placing the sensors on wheels (i.e. the cart or robot). We find that milliEgo works reasonably well when exposed to cart or robot motion which consists mainly of forward and backward movement down long hallways, but fails when exposed to sideways movement or fast rotations (Fig. \ref{fig:showcase_odom_b}). This is likely due to the sparse and fickle nature of radar point clouds.

\para{Robustness to Out-of-Distribution (OOD) Motions:} To what extent is \name's odometry pipeline invariant to platform-specific features? To answer this question, we train all odometry methods on the handheld trajectories in our dataset and test on the cart-driven and robot-driven parts of our dataset. As shown in Table \ref{tab:odometry_generalization}, \name\ is far more robust to shifts in platforms than other methods. This is due to IMU-aided methods leveraging and thus overfitting to platform-specific side-channel information (e.g., walking gaits, floor tile spacing, imperfections in wheels, etc. \cite{chen_ionet_2018, herath_ronin_2020, chen_rnin-vio_2021}). As expected, RNIN has the worst OOD performance with milliEgo performing slightly better due to access to radar data (Fig. \ref{fig:showcase_odom_c}). \name\ outperforms the baselines due to not leveraging such side channel information.

\subsection{SLAM Performance}

\begin{figure*}[t!]
    \centering
    \includegraphics[width=.8\linewidth]{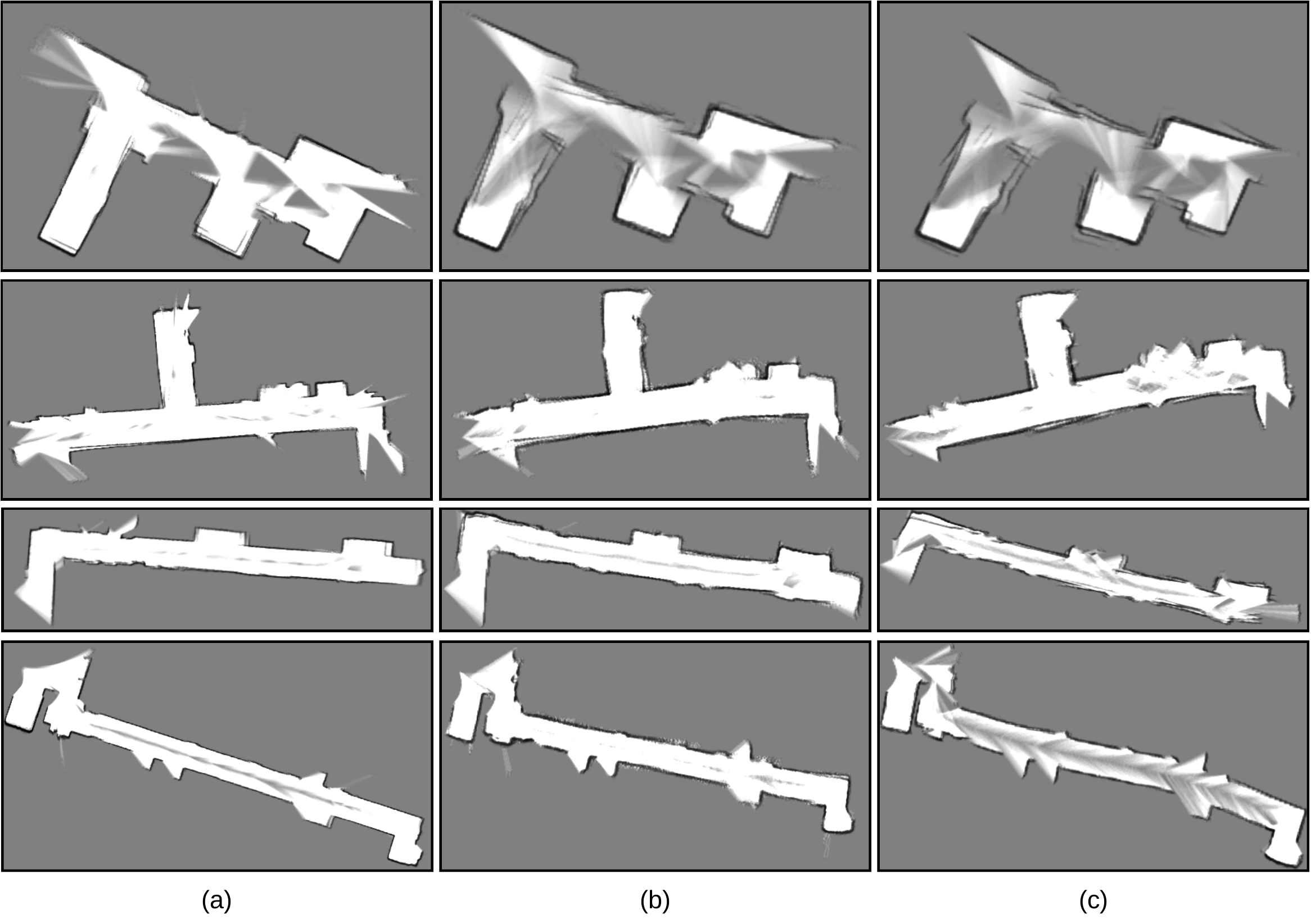}\vspace{-0.2in}
    \caption{\textbf{Comparing map fragments.} We compare rows 4 and 5 in Table \ref{tab:slam}. (a) Pseudo-ground truth map obtained from the stereo depth camera. (b) \name\ map reconstruction. (c) RadarHD map reconstruction. \name\ enables better mapping via artifact and multipath suppression.}
    \label{fig:showcase_map}
\end{figure*}

\begin{figure*}[t!]
    \centering
    \includegraphics[width=.85\linewidth]{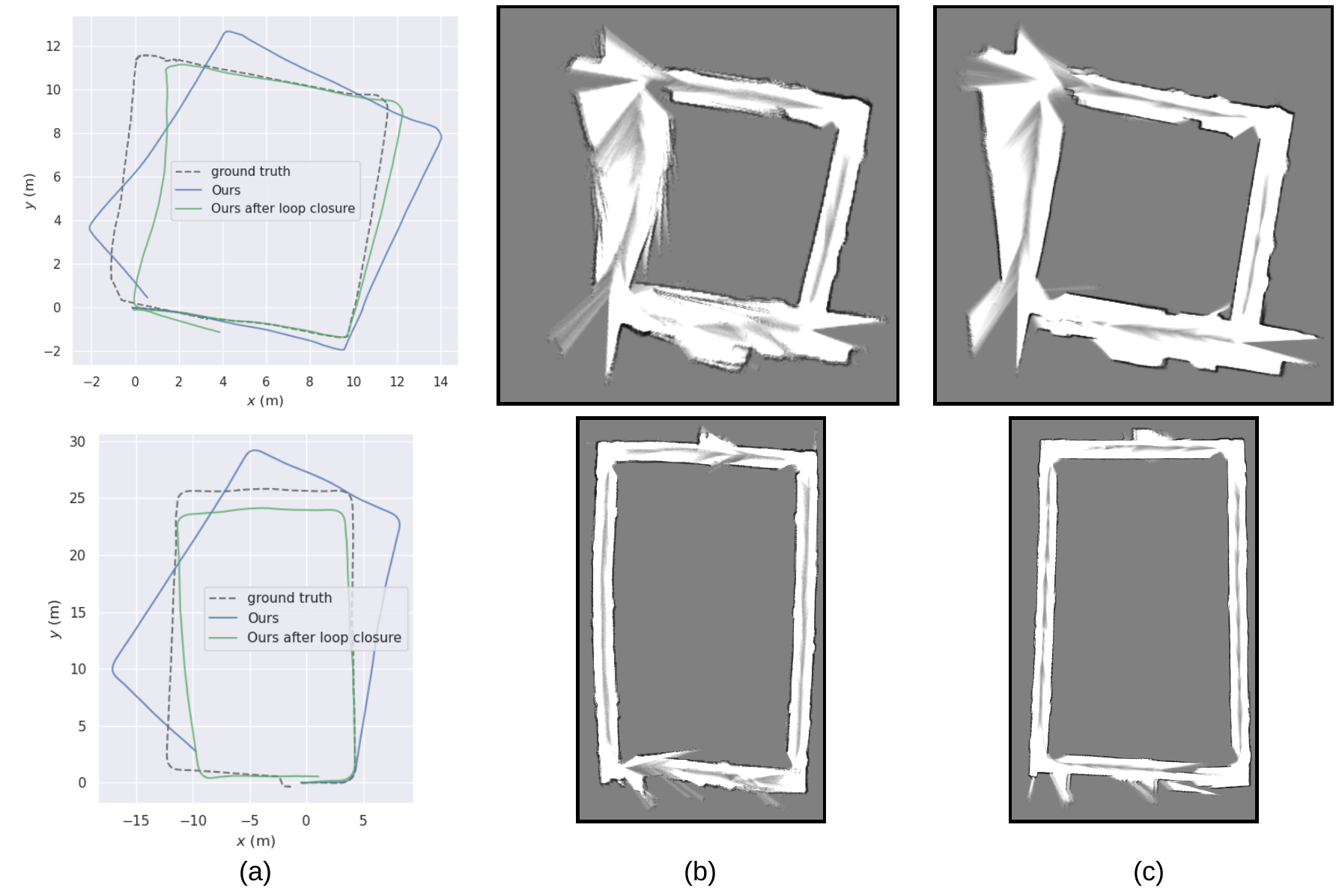}\vspace{-0.2in}
    \caption{\textbf{\name\ enables large-scale consistent maps.} \name's odometry has low drift over long distances, enabling long-range loop closures. (a) Trajectory estimate before and after loop closure. (b) Estimated map. (c) Pseudo ground-truth from depth + tracking camera.}
    \label{fig:showcase_slam}
\end{figure*}

Next, we evaluate the quality of trajectories and maps obtained after feeding odometry and scans from the baselines and our method into the SLAM backend \cite{cartographer}. The results are summarized in Table \ref{tab:slam}. We note that the performance of the scan matching only approach \cite{radarhd} is the worst, followed by methods leveraging odometry sources. This is due to the prevalence of degenerate environments in indoor environments such as long hallways where repeating features like flat walls confuse scan matching algorithms.  By contrast, \name's performance stems from a combination of its accurate odometry and scans, which the pose graph backend can optimize to yield high quality trajectories and consistent maps.

\vspace{3pt}\noindent\textbf{Map Quality:} We show qualitative examples of \name's local map fragments in Fig. \ref{fig:showcase_map}. We find that in spite of having more points per scan, RadarHD does not add any improvement over \name. This is due to the excess points in RadarHD contributing noise (e.g., due to not suppressing 3D to 2D conversion artifacts), which impedes the SLAM backend's ability to optimize and coherently combine scans into a consistent map.

\vspace{3pt}\noindent\textbf{Loop Closures:} The lack of compounding errors due to high-quality local mapping means that \name\ enables long range loop closures, such as those in excess of 50m. We show examples in Fig. \ref{fig:showcase_slam}, where the SLAM backend successfully closes long range loops.

\vspace{3pt}\noindent\textbf{Generalization Across Environments:} We test whether \name\ can successfully generalize across environments. We choose a subset of the 3 most distinctive buildings in our dataset (A,B,C). We test on trajectories in one building, and train on trajectories in the remaining buildings. The results are shown in Table \ref{tab:building}. We observe a small but minor performance decrease, which suggests that \name\ learns mostly environment-invariant features.

\begin{table}[t!]
\centering
\rowcolors{3}{gray!10}{}
\begin{tabular}{cccccc}
\toprule
\multicolumn{1}{c}{\multirow{2}{*}{\textbf{Building}}} & \multicolumn{2}{c}{\textbf{ATE}}                            & \multicolumn{2}{c}{\textbf{RE}}                             &  \\ \cmidrule{2-3} \cmidrule{4-5}
\multicolumn{1}{c}{}                                 & \multicolumn{1}{c}{Pos (m)} & \multicolumn{1}{c}{Rot (rad)} & \multicolumn{1}{c}{Pos (m)} & \multicolumn{1}{c}{Rot (rad)} &  \\ \midrule
A                               & 0.635                       & 0.102                         &  0.018                      & 0.100                         &  \\
B                               & 0.696                       & 0.179                         &  0.016                      & 0.175                         &  \\
C                               & 0.492                       & 0.105                         &  0.016                      & 0.104                         &  \\ \midrule
\textbf{Mean}                   & {0.608}              & {0.129}                &  {0.017}             & {0.126}                &
\\ \bottomrule
\end{tabular}
\caption{Testing \name's generalization across environments (tested on one building and trained on the rest).}
\label{tab:building}
\vspace{-0.2in}
\end{table}

\subsection{Computational Requirements}

\begin{table}[t!]
\centering
\rowcolors{2}{}{gray!10}
\begin{tabular}{lcccccc}
\toprule
\multirow{2}{*}{\textbf{Device}} & \multicolumn{5}{c}{\textbf{Mean Runtime} (ms)} & \multirow{2}{*}{\textbf{Total}} \\ 
\cmidrule{2-6} 
                    & Preproc.         & Trans. & Rot.     & Map    & SLAM  & \\ \hline
Desktop             & 9.53             & 2.23   & 6.60     & 3.11   & 0.93  & 17.06 \\ 
AGX Orin            & 11.76            & 7.57   & 14.78    & 9.43   & 4.68  & 31.22 \\ \hline
\end{tabular}
\caption{Mean runtime of each stage of \name's pipeline on different devices. The total runtime is computed along the longest path.}
\label{tab:runtime_profiling}
\vspace{-0.2in}
\end{table}

Finally, we quantify \name's computational requirements. We deploy \name\ on two different host devices (a desktop computer and single-board edge computer). The desktop computer is equipped with a 12-core AMD Ryzen 9 3900X processor, 32 GB RAM, and an NVIDIA RTX 3090 GPU. The Jetson AGX Orin \cite{jetson_agx_orin} is a small form-factor computer geared towards mobile edge machine learning applications such as robotics. It is equipped with a 12-core Arm Cortex-A78AE CPU, 32 GB RAM, and an 2048-core NVIDIA Ampere architecture GPU with 64 Tensor Cores.

We report the running time of each of \name's components in Table \ref{tab:runtime_profiling}. To determine the total pipeline runtime, we sum along the longest (bottleneck) path. We obtain a total runtime of 17.06 ms (desktop) and 31.22 ms (AGX Orin). This translates to an update rate of 59 Hz and 32 Hz, which is corresponds to super-realtime and realtime performance respectively. We conclude that \name\ is computationally efficient enough to be deployed as a realtime system even on edge compute devices such as the AGX Orin.

\section{Related Work}
\label{sec:related}

\noindent\textbf{Methods Using Large Radars: }Large spinning radars have been used to perform localization and mapping on vehicles in outdoor environments. Recent work in this space includes \cite{cen_precise_2018}, RADARODO \cite{monaco_radarodo_2020}. Pharao \cite{park_pharao_2020}, RadarSLAM \cite{hong_radarslam_2020}. As GPS is available in outdoor environments, some of these approaches fuse their estimates using GPS \cite{liang_scalable_2020}. \name\ differs along two axes. First, these methods are suited for larger and more expensive radars with considerably greater resolution. Second, these works do not address the unique challenges of using radar within indoor environments such as multipath effects and phantom reflections.

\vspace{3pt}\noindent\textbf{Methods Using Commodity Hardware:} As large radars are impractical indoors and on resource constrained edge devices, there has been recent interest in leveraging small low cost radars and commodity-grade IMUs. Works involving fusion of radar and IMU data includes \cite{kramer_radar-inertial_2020} which relies on classical methods and milliEgo \cite{lu_milliego_2020} which leverages deep learning. We note there is also a body of work focusing purely on IMU-based odometry, such as RONIN \cite{herath_ronin_2020} RNIN-VIO \cite{chen_rnin-vio_2021}, which could be fused with radar in a loosely-coupled framework. Unlike these works, \name\ develops purely radar-based odometry by relying on doppler shift, and does not require inertial sensors. Finally, there are works also using doppler shift for egomotion estimation such as \cite{kwon_radar_2023} and BatMobility\cite{sie2023batmobility}. BatMobility leverages doppler shift and machine learning to estimate ground-parallel translational motion of an autonomous UAV. \cite{kwon_radar_2023} uses doppler shift to estimate translation and line-fitting with RANSAC to estimate rotation within a hallway environment. In contrast to these methods, \name\ does not require strong apriori assumptions about the environment (e.g. existence of a flat ground surface in BatMobility, or a hallway environment with parallel flat walls as in \cite{kwon_radar_2023}).

\vspace{3pt}\noindent\textbf{Methods Based on Super-Resolution:} Radar is noted for its low resolution when compared with its optical counterpart (i.e. lidar). %
Several works attempt to solve this issue through deep-learning based super resolution techniques. milliMap \cite{millimap} applies GANs to densify top-down projections of sparse radar point clouds. RadarHD \cite{radarhd} applies an assymetric UNet segmentation to upsample range-azimuth heatmaps across the angular dimension. Unlike these works, we identify 3D to 2D conversion from ceilings and floors as a common source of mapping artifacts in indoor environments, and address this using our preprocessing algorithm which takes into account elevation. Finally, while milliMap relies on explicit ground truth inputs for odometry and RadarHD infers them using scan matching on radar point clouds, \name\ relies on doppler-based odometry. %

\section{Concluding Discussion}
\label{sec:conclusion}

We present \name, a radar-native SLAM pipeline using only a small commodity mmWave radar. \name\ exploits the properties of small FMCW radars in indoor environments to simultaneously perform accurate odometry and mapping. \name\ is flexible enough to be mounted on different platforms, such as headsets or handsets mounted on humans or robots close to the ground. We conclude with a discussion of limitations and future work:

\vspace{3pt}\noindent\textbf{Range Limits:} We use a commodity-grade stereo depth camera with an ideal operating range of of less than 5m. As such, we set the radar's maximum range to match that of the depth camera. Although this suffices for most indoor environments, it might be prudent to extend the range of radar for operation in larger spaces or in outdoor scenarios. This will require retraining \name's models using ground truth from a more expensive lidar sensor.

\vspace{3pt}\noindent\textbf{Speed Limits:} As explained in Sec. \ref{sec:implementation}, our dataset is collected by humans carrying a sensor either by hand, on a cart, or on a slow moving ground robot. Hence, the maximum linear speed in our dataset is 1.6 m/s. We set our radar parameters such that the maximum unambiguous velocity corresponds to around the same limit. We believe this is a reasonable speed for most ground robots and even humans. However, if a larger speed limit is necessary, it is possible to increase the maximum unambiguous velocity of the radar by reconfiguring the chirp parameters.

\vspace{3pt}\noindent\textbf{Dynamic Objects:} As mentioned in Sec. \ref{sec:implementation}, we do not train or test our system in the presence of a dynamic objects in the environment. Similar to how the performance of camera or lidar based SLAM deteriorates in dynamic environments \cite{minoda_viode_2021}, the presence of moving objects within the environment is likely to adversely affect the performance of the system. We leave the resolution of such issues to future work.

\begin{acks}
We thank the anonymous reviewers and our shepherd Wen Hu for their insightful comments and suggestions on improving this paper. This work was supported in part by NSF RINGS Award 2148583. 
\end{acks}

\bibliographystyle{acm}
\bibliography{sample}

\end{document}